\documentclass[11pt,a4paper]{article}
\usepackage[hyperref]{emnlp2020}
\usepackage{times}
\usepackage{latexsym}

\usepackage{microtype}
\usepackage{multirow}
\usepackage{comment}
\usepackage{graphicx}
\usepackage{amsmath}
\aclfinalcopy 
\usepackage{microtype}

\def\aclpaperid{645} 

\title{Multi-View Sequence-to-Sequence Models with Conversational Structure for Abstractive Dialogue Summarization}

\author{Jiaao Chen \\
  School of Interactive Computing \\
  Georgia Institute of Technology \\
  \texttt{jiaaochen@gatech.edu} \\\And
  Diyi Yang \\
  School of Interactive Computing \\
  Georgia Institute of Technology \\
  \texttt{dyang888@gatech.edu} \\}

\date{}

\begin{document}
\maketitle
\begin{abstract}
Text summarization is one of the most challenging and interesting problems in NLP.
Although much attention has been paid to summarizing structured text like news reports or encyclopedia articles, summarizing conversations---an essential part of human-human/machine interaction where most important pieces of information are scattered across various utterances of different speakers---remains relatively under-investigated. 
This work proposes a multi-view sequence-to-sequence model by first extracting conversational structures of unstructured daily chats from different views to represent conversations and then utilizing a multi-view decoder to incorporate different views to generate dialogue summaries. Experiments on a large-scale dialogue summarization corpus
demonstrated that our methods significantly outperformed previous state-of-the-art models via both automatic evaluations and human judgment. We also discussed specific challenges that current approaches faced with this task.  We
have publicly released our code at \url{https://github.com/GT-SALT/Multi-View-Seq2Seq}.
\end{abstract}

\section{Introduction}
\begin{table*}[]
\centering
\begin{tabular}{|rl||l||l|}
\hline
\multicolumn{2}{|l||}{\textbf{Conversation}}             & \textbf{Topic View}                     & \textbf{Stage View}              \\ \hline
James:  & Hey! I have been thinking about you : )        & \multirow{2}{*}{Greetings}         & \multirow{3}{*}{Openings}   \\ \cline{1-2}
Hannah: & Oh, that's nice ; )                            &                                    &                             \\ \cline{1-3}
James:  & What are you up to?                            & \multirow{3}{*}{Today's plan}      &                             \\ \cline{1-2} \cline{4-4} 
Hannah: & I'm about to sleep                             &                                    & \multirow{2}{*}{Intention}  \\ \cline{1-2}
James:  & I miss u. I was hoping to see you              &                                    &                             \\ \hline
Hannah: & Have to get up early for work tomorrow         & \multirow{3}{*}{Plan for tomorrow} & \multirow{6}{*}{Discussion} \\ \cline{1-2}
James:  & What about tomorrow?                           &                                    &                             \\ \cline{1-2}
Hannah: & To be honest I have plans for tomorrow evening &                                    &                             \\ \cline{1-3}
James:  & Oh ok. What about Sat then?                    & \multirow{2}{*}{Plan for Saturday} &                             \\ \cline{1-2}
Hannah: & Yeah. Sure I am available on Sat               &                                    &                             \\ \cline{1-3}
James:  & I'll pick you up at 8?                         & \multirow{2}{*}{Pick up time}      &                             \\ \cline{1-2} \cline{4-4} 
Hannah: & Sounds good. See you then.                     &                                    & Conclusion                  \\ \hline \hline
\multicolumn{1}{|l}{\textbf{Summary}}  & \multicolumn{3}{l|}{James misses Hannah. They agree for James to pick Hannah up on Saturday at 8.} \\ \hline 
\end{tabular}
\caption{Example conversation from SAMSum \cite{gliwa2019samsum} with its topic view and stage view (extracted by our methods), and the human annotated summary.}\label{tbl:example}
\end{table*}

We live in an information age where communications between human and human/machine are increasing exponentially in the form of textual dialogues between users and users-agents \cite{kester2004conversation}.
It is challenging and time-consuming to review all the content before starting any conversations especially when the chatting history becomes very long \cite{gao2020standard}. How to process and organize those interaction activities into concise and structured data, i.e. conversation summarization, becomes technically and socially important. 

Most existing research efforts on text summarization have been focused on single-speaker documents like news reports \cite{nallapati-etal-2016-abstractive, See_2017}, scientific publications \cite{DBLP:journals/corr/abs-1804-08875} or encyclopedia articles \cite{DBLP:journals/corr/abs-1801-10198}, where structured text is usually used to elaborate a core idea in the third-person point of view, and the information flow is very clear through paragraphs or sections. Different from these structured documents, conversations are often informal, verbose and repetitive, sprinkled with false-starts, back channeling, reconfirmations, hesitations, speaker interruptions \cite{sacks1978simplest} and the salient information is scattered in the whole chat, making current summarization models hard to focus on many informative utterances. Take the conversation in Table~\ref{tbl:example} as an example, turns, informal words, abbreviations, and emoticons all introduce new forms of challenges to the task of summarization. This calls for the design and development of new methods for dialogue summarization instead of directly applying current document summarization models.

There has been some recent research on conversation summarization such as directly deploying existing document summarization models \cite{gliwa2019samsum} and exploring multi-sentence compression \cite{shang-etal-2018-unsupervised}, however, most of them haven't utilized specific conversational structures, which refer to the way utterances are organized in order to make the conversation meaningful, enjoyable and understandable \cite{sacks1978simplest}, in dialogues -- a key factor that differentiates dialogues from structured documents.  
As a way of using language socially of ``doing things with words'' together with other persons, the conversation has its own dynamic structures that organize utterances in certain orders to make the conversation meaningful, enjoyable, and understandable \cite{sacks1978simplest}.
Although there are a few exceptions such as utilizing topic segmentation \cite{Liu_2019, li-etal-2019-keep}, dialogue acts \cite{Goo_2018} or key point sequence \cite{10.1145/3292500.3330683}, they either need extensive expert annotations of discourse acts\cite{Goo_2018, 10.1145/3292500.3330683}, or only encode conversations based on their topics \cite{Liu_2019}, which fails to capture rich conversation structures in dialogues. 

Even one single conversation can be viewed from different perspectives, resulting in multiple conversational or discourse patterns. For instance, in Table~\ref{tbl:example},
based on what topics were discussed (\textbf{topic view}) \cite{galley-etal-2003-discourse, Liu_2019, li-etal-2019-keep},
it can be segmented into \emph{greetings}, \emph{today's plan}, \emph{plan for tomorrow}, \emph{plan for Saturday} and \emph{pick up time}; from a conversation progression perspective (\textbf{stage view}) \cite{ritter-etal-2010-unsupervised, paul-2012-mixed, althoff-etal-2016-large}, the same dialogue can be categorized into \emph{openings}, \emph{intention}, \emph{discussion}, and \emph{conclusion}. 
From a coarse perspective (\textbf{global view}), conversations can be treated as a whole, or each utterance can serve as one segment (\textbf{discrete view}). Models that only utilized a fixed topic view of the conversation \cite{joty-etal-2010-exploiting, Liu_2019} may fail to capture its comprehensive and nuanced conversational structures, and any amount of information loss introduced by the conversation encoder may lead to larger error cascade in the decoding stage. 
To fill these gaps, we propose to combine those multiple, diverse views of conversations in order to generate more precise summaries. 

To sum up, our contributions are: (1) we propose to utilize rich conversational structures, i.e., structured views (topic view and stage view) and the generic views (global view and discrete view) for abstractive conversation summarization.
(2) We design a multi-view sequence-to-sequence model that consists of a conversation encoder to encode different views and a multi-view decoder with multi-view attention to generate dialogue summaries.
(3) We perform experiments on a large-scale conversation summarization dataset, SAMSum \cite{gliwa2019samsum}, and demonstrate the effectiveness of our proposed methods.
(4) We conduct thorough error analyses and discuss specific challenges that current approaches faced with this task. 
\begin{figure*}[t]
\centering
\includegraphics[width=2.1\columnwidth]{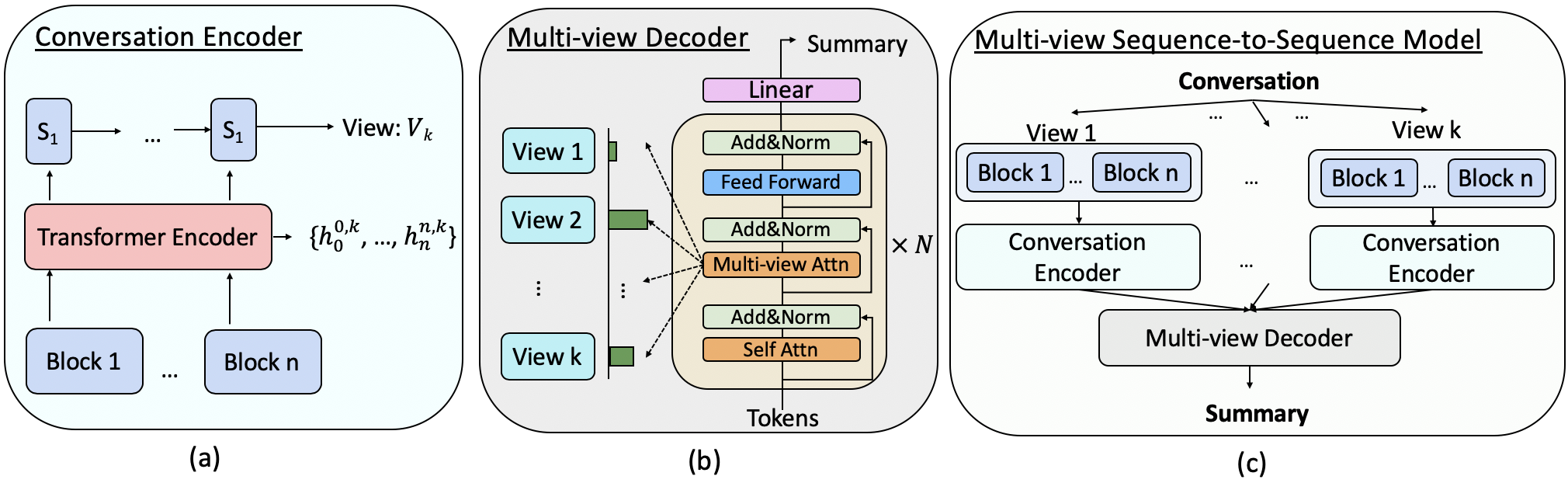}
\caption{Model architecture. Different views of conversations are first extracted automatically, and then encoded through the conversation encoder (a) and combined in the multi-view decoder to generate summaries (b). In the conversation encoder, each view (consists of blocks) is encoded separately and the block's representations $S_i$ are encoded through LSTM to represent the view. In the multi-view decoder, the model decides attention weights over different views and then attend to each token in different views through the multi-view attention. }
\label{Fig:model}
\end{figure*}
\section{Related Work}
\paragraph{Document Summarization}
Document summarization has received extensive research attention, especially for abstractive summarization. For instance, \citet{rush-etal-2015-neural} introduced to use sequence-to-sequence models for abstractive text summarization. \citet{See_2017} proposed a pointer-generator network to allow copying words from the source text to handle the OOV issue and avoid generating repeated content. \citet{paulus2018a, DBLP:journals/corr/abs-1805-11080} further utilized reinforcement learning to select the correct content needed by summarization. 
Large-scale pre-trained language models \cite{Liu_2019_pretrained, raffel2019exploring, lewis2019bart} have also been introduced to further improve the summarization performance. Other line of work explored long-document summarization by utilizing discourse structures in text \cite{cohan-etal-2018-discourse}, introducing hierarchical models \cite{Fabbri_2019} or modifying attention mechanisms \cite{beltagy2020longformer}. There are also recent studies looking at the faithfulness in document summarization \cite{Cao2018FaithfulTT, zhu2020boosting}, in order to enhance the information consistency between summaries and the input.

\paragraph{Dialogue Summarization}
When it comes to the summarization of dialogues, \citet{shang-etal-2018-unsupervised} proposed a simple multi-sentence compression technique to summarize meetings. \citet{10.1145/3308558.3313619, zhu2020endtoend} introduced turn-based hierarchical models that encoded each turn of utterance first and then used the aggregated representation to generate summaries. 
A few studies have also paid attention to utilizing conversational analysis for generating dialogue summaries, such as leveraging dialogue acts \cite{Goo_2018}, key point sequence \cite{10.1145/3292500.3330683} or topics \cite{Liu_2019, li-etal-2019-keep}. 
However, they either needed a large amount of human annotation for dialogue acts, key points or visual focus \cite{Goo_2018, 10.1145/3292500.3330683, li-etal-2019-keep},
or only utilized topical information in conversations \cite{li-etal-2019-keep, Liu_2019}.

These prior work also largely ignored diverse conversational structures in dialogues, for instance, reply relations among participants \cite{mayfield-etal-2012-hierarchical, zhu2019did}, dialogue acts \cite{ritter-etal-2010-unsupervised, paul-2012-mixed}, and conversation stages \cite{althoff-etal-2016-large}.  Models that only utilized a fixed topic view of the conversation \cite{galley-etal-2003-discourse, joty-etal-2010-exploiting} may fail to capture its comprehensive and nuanced conversational structures, and any amount of information loss introduced by the conversation encoder may lead to larger error cascade in the decoding stage. 
To fill these gaps, we propose to leverage diverse conversational structures including topic segments, conversational stages, dialogue overview, and utterances to design a multi-view model for dialogue summarization.

\section{Method}
Conversations can be interpreted from different views and every single view enables the model to focus a specific aspect of the conversation. To take advantages of those rich conversation views, we design a Multi-view Sequence-to-Sequence Model (see Figure~\ref{Fig:model}) that firstly extracts different views of conversations (Section~\ref{Sec:view_extraction}) and then encodes them to generate summaries (Section~\ref{Sec:s2smodel}).

\subsection{Conversation View Extraction} \label{Sec:view_extraction}
Conversation summarization models may easily stray among all sorts of information across various speakers and utterances especially when conversations become long. Naturally, if informative structures in the form of small blocks can be explicitly extracted from long conversations, models may be able to understand them better in a more organized way.
Thus, we first extract different views of structures from conversations.

\paragraph{Topic View} Although conversations are often less structured than documents, they are mostly organized around topics in a coarse-grained structure \cite{honneth1988social}. For instance, a telephone chat could possess a pattern of ``\textsl{greetings $\rightarrow$  invitation $\rightarrow$ party details $\rightarrow$ rejection}'' from a topical perspective. Such explicit view and topic flow could help models interpret conversations more precisely and generate summaries that cover important topics. Here we combine the classic topic segment algorithm, C99 \cite{choi-2000-advances} that segments conversations based on inter-sentence similarities, with recent advanced sentence representations Sentence-BERT \cite{Reimers_2019}, to extract the topic view. Specifically, each utterance $u_i$ in a conversation $\mathbf{C} = \{u_1, u_2, ..., u_m\}$ is first encoded into hidden vectors via Sentence-BERT. Then the conversation $\mathbf{C}$ is divided into blocks $\mathbf{C}_{topic} = \{\mathbf{b}_1, ..., \mathbf{b}_n\}$ through C99, where $\mathbf{b}_i$ is one block that contains several consecutive utterances, such as the topic view described in Table~\ref{tbl:example}.

\begin{figure}[t]
\centering
\includegraphics[width=0.8\columnwidth]{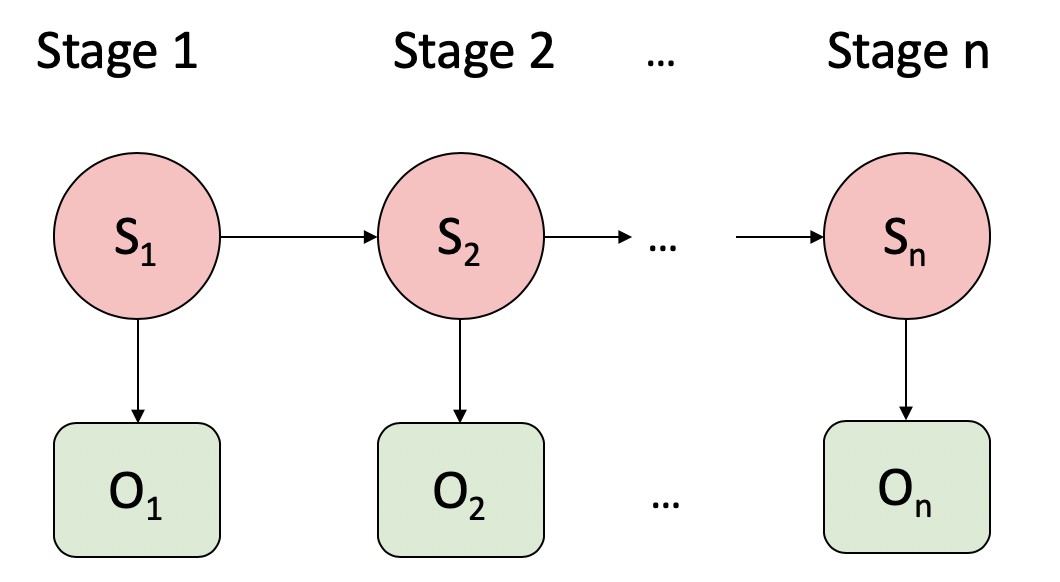}
\caption{Allowed state transitions for the HMM conversation
model. $S_i$ are conversation stages, $O_i$ are sentences' encoded representations. Conversation stages evolve in an increasing order from 1 to $n$.}\label{Fig:hmm_model}
\end{figure}
\paragraph{Stage View} As a way of doing things with words socially together with other people, conversation organizes utterances in certain orders to make it meaningful, enjoyable, and understandable. \cite{sacks1978simplest, althoff-etal-2016-large} For example, counseling conversations are found to follow a common pattern of ``\textsl{introductions $\rightarrow$  problem exploration $\rightarrow$ problem solving $\rightarrow$ wrap up}'' \cite{althoff-etal-2016-large}. Such conversation stage view provides high-level sketches about the functions or goals of different parts in conversations, which could help models focus on the stages with key information.

We follow \citet{althoff-etal-2016-large} to extract stages through a Hidden Markov Model (HMM). We impose a fixed ordering
on the stages and only allow transitions from the current stage to the next one. 
The observations in the HMM model are the encoded representations $h_i$ from Sentence-BERT. 
We set the number of hidden stages as 4. 
Similar to the topic view extraction, we segment the conversations into blocks $\mathbf{C}_{stage} = \{\mathbf{b}_1, ..., \mathbf{b}_n\}$, where $\mathbf{s}_i$ is one block that contains several consecutive utterances. 
We interpret the inferred stages qualitatively and further visualize the top 6 frequent words appearing in each stage in Table~\ref{Tab:hmm_interperation}. We found that conversations around daily chats usually start with \textsl{openings}, introduce the goals/focus of the conversation followed by discussions of the details, and finally conclude with certain endings.  Table~\ref{tbl:example} shows an example of the stage view.

\begin{table}[t]
\centering
\begin{tabular}{|c|c|l|}
\hline
\textbf{Stage} & \textbf{Interpretation} & \textbf{Top Freq Words}                       \\ \hline
1      & Openings      &\begin{tabular}[l]{@{}l@{}}  hey, hi, good, \\ yeah,going, time      \end{tabular} \\ \hline
2      & Intentions      &\begin{tabular}[l]{@{}l@{}}  need, like, think, \\ get, want, really \end{tabular} \\ \hline
3      & Discussions     &\begin{tabular}[l]{@{}l@{}}  will, know, time, \\come,tomorrow, meet \end{tabular} \\ \hline
4      & Conclusions     &\begin{tabular}[l]{@{}l@{}}  thanks, ok, see, \\great, thank, sure \end{tabular} \\ \hline
\end{tabular} \caption{The top 6 frequent words appearing in each stage and the interpretations for different stages.} \label{Tab:hmm_interperation}
\end{table}

\paragraph{Global View and Discrete View} In addition to the aforementioned two structured views, conversations can also be naturally viewed from a relatively coarse perspective, i.e., a global view that concatenates all utterances into one giant block \cite{gliwa2019samsum}, and a discrete view that separates each utterance into a distinct block \cite{ liu-chen-2019-reading,gliwa2019samsum}.

\subsection{Multi-view Sequence-to-Sequence Model} \label{Sec:s2smodel}
We extend generic sequence-to-sequence models to encode and combine different conversation views. To better utilize semantic information in recent pre-trained models, we implement our base encoders and decoders with a transformer based pre-trained model, BART \cite{lewis2019bart}. Note that our multi-view sequence-to-sequence model is agnostic to BART with which it is initialized.

\paragraph{Conversation Encoder} 
Given a conversation under a specific view $k$ with $n$ blocks: $\mathbf{C}_k  = \{\mathbf{b}_1^k, ..., \mathbf{b}_n^k\}$, each token $x_{i,j}^k$ in a block $\mathbf{b}_j^k = \{x_{0,j}^{k}, x_{1,j}^{k}, ..., x_{m,j}^{k}\}$ is first encoded through the conversation encoder $\mathbf{E}$ , e.g., BART encoder as shown in Figure 1(a), into hidden representations:
\begin{equation}
    \{h_{0,j}^{k}, h_{1,j}^{k}, ..., h_{m,j}^{k}\} = \mathbf{E}(\{x_{0,j}^{k}, x_{1,j}^{k}, ..., x_{m,j}^{k}\})
\end{equation}
Note that we add special tokens $x_{0,j}^{k}$ at the beginning of each block and use these tokens' representations to describe each block, i.e., $S_j^k = h_{0,j}^{k}$. 

To depict different views using hidden vectors, we aggregate the information from all blocks in one conversation through LSTM layers \cite{10.1162/neco.1997.9.8.1735}: 
\begin{equation}
    S_j^k = \text{LSTM}(h_0^{j,k}, S_{j-1}^k), j \in [1, n]
\end{equation}
We use the last hidden state $S_n^k$ to represent the current view $k$, denoted as $V_k$.

\paragraph{Multi-view Decoder}
Different views could provide different types of conversational aspects for models to learn and further determine which set of utterances should deserve more attention in order to generate better dialogue summaries. As a result, the ability to strategically combine different views is essential. To this end, we propose a transformer based multi-view decoder to integrate encoded representations from different views and generate summaries as shown in Figure 1(b). 

The input to the decoder contains $l-1$ previously generated tokens $t_1, ..., t_{l-1}$. Via our multi-view decoder $\mathbf{D}$, the $l$-th token is predicted via:
\begin{align}
    \{y_1, ..., y_{l-1}\} = \mathbf{D}&(\{t_1, ..., t_{l-1}\}, \mathbf{E}(C)) \\
    P(\tilde{t}_l|t_{<l}, \mathbf{C}) &= \text{Softmax}(W_p y_{l-1})
\end{align} 
Here, $W_p$ is a parameter to be learned.

Different from generic transformer decoder, we introduce a multi-view attention layer in each transformer block. Multi-view attention layer first decides the importance $\alpha_{k}$ of each view $V_k$ through:
\begin{align}
u_{k} &=\tanh \left(W V_{k}+b\right) \\ 
\alpha_{k} &=\frac{\exp \left(u_{k}^{\top} v\right)}{\sum_{i} \exp \left(u_{i}^{\top} v\right)}  
\end{align} \\
where $v$ is a randomly initialized context vector; $W$ and $b$ are parameters. To avoid the attention weights being too similar to each other as views are actually encoded from a similar context, we utilize a sharpening function over $\alpha_{k}$ with a temperature T: $\tilde{\alpha}_{k} = {\alpha_{k}^{\frac{1}{T}}}/{\sum_{i} \alpha_{i}^{\frac{1}{T}}}$. When $T \rightarrow 0$, the attention weights will behave like a one-hot vector. 

Then the multi-head attention is performed over conversation tokens $h_{i,j}^{k}$ from different views $k$ and form $A^k$ separately. The attended results are further combined based on the view-attention weights $\tilde{\alpha}_{k}$ and continue forward passing:
\begin{align}
    \tilde{A} = \sum_k \tilde{\alpha}_{k} A^k
\end{align}

\paragraph{Training} We minimize the cross entropy loss during training: 
\begin{align}
    L = - \sum \log P(\tilde{t}_l|t_{<l}, \mathbf{C}) 
\end{align}
Specifically, we apply the teacher forcing strategy: at training time, the inputs are previous tokens from the ground truth; at test time, the inputs are previous tokens predicted by the decoder.

\section{Experiments}
\subsection{Dataset and Baselines}
We evaluate our model on a large-scale dialogue summary dataset SAMSum \cite{gliwa2019samsum} that has 14732 dialogues with human-written summaries. The data statistics are shown in Table~\ref{Tab: data_statistics}. SAMSum contains messenger-like conversations about daily topics, such as chit-chats, arranging meetings, discussing events, etc. 
We compare our Multi-view Sequence-to-Sequence Model (Multi-view BART) with several baseline models:

\begin{table*}[t]
\centering
\begin{tabular}{|rl|ccc|ccc|ccc|} 
\hline
\multicolumn{2}{|r|}{\multirow{2}{*}{\textbf{\# Conversations}}} & \multicolumn{3}{c|}{\textbf{\# Participants}} & \multicolumn{3}{c|}{\textbf{\# Turns}} & \multicolumn{3}{c|}{\textbf{Reference Length}} \\ \cline{3-11} 
                        &                              & Mean     & Std     & Interval    & Mean    & Std   & Interval  & Mean    & Std     & Interval  \\ \hline
Train                   & 14732                        & 2.40     & 0.83    & [1, 14]          & 11.17   & 6.45  & [1, 46]      & 23.44   & 12.72   & [2, 73]     \\
Dev                     & 818                          & 2.39     & 0.84    & [2, 12]          & 10.83   & 6.37  & [3, 30]       & 23.42   & 12.71   & [4, 68]      \\
Test                    & 819                          & 2.36     & 0.83    & [2, 11]         & 11.25   & 6.35  & [3, 30]   & 23.12   & 12.20   & [4, 71]      \\ \hline 
\end{tabular} \caption{SAMSum dataset statistics. \textsl{Interval} denotes the minimum and maximum range.} \label{Tab: data_statistics}
\end{table*}

\begin{itemize}
    \item \textbf{Pointer Generator} \cite{See_2017}: Following \citet{gliwa2019samsum}, we added separators between each utterance (\textbf{discrete view}) and used it as input for pointer generator model.
    \item \textbf{DynamicConv + GPT-2/News} \cite{DBLP:journals/corr/abs-1901-10430}: We followed \citet{gliwa2019samsum} to use GPT-2 to initialize token embeddings \cite{radford2019language}. We also added news summarization corpus CNN/DM \cite{nallapati-etal-2016-abstractive} as extra training data. 
    \item \textbf{Fast Abs RL Enhanced} \cite{DBLP:journals/corr/abs-1805-11080} first selects salient sentences and then rewrites them abstractively via sentence-level policy gradient methods. We combined it with the \textbf{global view} \cite{gliwa2019samsum}. 
    
    \item \textbf{BART + Generic views} \cite{lewis2019bart}  utilized BART, a denoising autoencoder for pretraining sequence-to-sequence models, together with generic views (\textbf{global view} and \textbf{discrete view}). We used the BART-large model with its default settings \footnote{\url{https://github.com/pytorch/fairseq}}.
\end{itemize}

\begin{table*}[t]
\centering
\small
\begin{tabular}{|c|c|ccc|ccc|ccc|}
\hline
\multirow{2}{*}{\textbf{Model}}   & \multirow{2}{*}{\textbf{Views}}          & \multicolumn{3}{c|}{\textbf{ ROUGE-1}} & \multicolumn{3}{c|}{\textbf{ ROUGE-2}} & \multicolumn{3}{c|}{\textbf{ ROUGE-L}}\\ \cline{3-11}
                                 &{}   & F       & P       & R       & F       & P       & R       & F       & P       & R       \\ \hline
Pointer Generator     & Discrete               & 0.401   & -       & -       & 0.153   & -       & -       & 0.366   & -       & -       \\
DynamicConv + GPT-2      & Global         & 0.418   & -       & -       & 0.164   & -       & -       & 0.376   & -       & -       \\
Fast Abs RL Enhanced & Global & 0.420   & -       & -       & 0.181   & -       & -       & 0.392   & -       & -  \\ 
DynamicConv + News & Discrete   & 0.454   & -       & -       & 0.206   & -       & -       & 0.415   & -       & -       \\  \hline
\multirow{2}{*}{BART}   & Discrete                 & 0.481   & 0.452   & 0.526   & 0.245   & 0.236   & 0.282   & 0.451   & 0.432   & 0.521   \\
{} & Global                   & 0.482   & 0.493   & 0.517   & 0.245   & 0.251   & 0.264   & 0.466   & 0.475   & 0.495   \\ \hline \hline
\multirow{2}{*}{BART$\dag$} & Stage                    & 0.487   & 0.483   & 0.540   & 0.251   & 0.248   & 0.282   & 0.472   & 0.469   & 0.515   \\
{} & Topic                    & 0.488   & 0.479   & 0.547   & 0.251   & 0.248   & 0.284   & 0.474   & 0.483   & 0.501   \\ \hline
\multirow{3}{*}{Multi-view BART$\dag$ }       &  Global + Stage               & 0.488  & 0.476   & 0.548  &0.251   & 0.246  & 0.285   & 0.472   & 0.462   & 0.521  \\  {} &  Global + Topic              & 0.488  & 0.488  & 0.535  & 0.251   & 0.252   & 0.275   & 0.473   & 0.474   & 0.509  \\ 
{}    & Topic + Stage                & \textbf{0.493}   & 0.511   & 0.522   & \textbf{0.256}   & 0.265   & 0.274   & \textbf{0.477}   & 0.493   & 0.499  \\ \hline 
\end{tabular} 
\caption{ ROUGE-1,  ROUGE-2 and  ROUGE-L scores for different models on the test set. Results are averaged over three runs. $\dag$ meant our methods or utilized views introduced by us. } \label{Tab: Main_results}
\end{table*}
\subsection{Model Settings\footnote{More details are shown in Section A in the Appendix.}}
We loaded the pre-trained ``bert-base-nli-stsb-mean-tokens''\footnote{\url{https://github.com/UKPLab/sentence-transformers}} for sentence-BERT to get representations for each utterance. For extracting the \textsl{topic view} via C99, we set the window size 4 and std coefficient 1. For extracting the \textsl{stage view}, we set the number of hidden states 4 in HMM. These hyper-parameters were set with a grid search.  The \textbf{BART + Structured views} (stage and topic views) used the same set of parameters as \textbf{BART + Generic views}. 
For \textbf{Multi-View BART}, we experimented with different view combinations: (1) the best generic view - global view, was combined with two structured views (stage and topic view) separately; (2) the best two structured views are also combined (topic + stage). The settings for BART encoder/decoder kept identical as baselines. We used a one-layer LSTM for encoding sections. The learning rate for section encoder and multi-view attention was set 3e-3. The temperature $T$ was 0.2. The beam search size during inference for all the models was 4.

\subsection{Results}
\textbf{Quantitative Results} 
We evaluated models with the standard metric  ROUGE Score (with stemming) \cite{lin2004automatic}, and reported  ROUGE-1,  ROUGE-2 and  ROUGE-L\footnote{Here we followed BART and used \url{https://github.com/pltrdy/rouge}. Note that different tools may generate different ROUGE scores.}.
Results on the test set for different models were shown in Table~\ref{Tab: Main_results}. Compared to \textsl{Pointer Generator}, using reinforcement learning to select important sentences first (\textsl{Fast Abs RL Enhanced}) slightly increased F scores. Adding pre-trained embeddings or extra documents training data to lightweight convolution models, (\textsl{DynamicConv + GPT-2/News}) lead to even better  ROUGE scores.  When using pre-trained transformer based model BART with generic views, all  ROUGE scores improved significantly, and \textsl{BART + Global} outperformed \textsl{BART + Discrete} especially in terms of  ROUGE-L F scores. 
Segmenting conversations into blocks from structured views (stage view and topic view) further boosted the performance, suggesting that our extracted conversation structures help conversational encoders to capture nuanced and informative aspects of dialogs.

\begin{figure}[t]
\centering
\includegraphics[width=0.95\columnwidth]{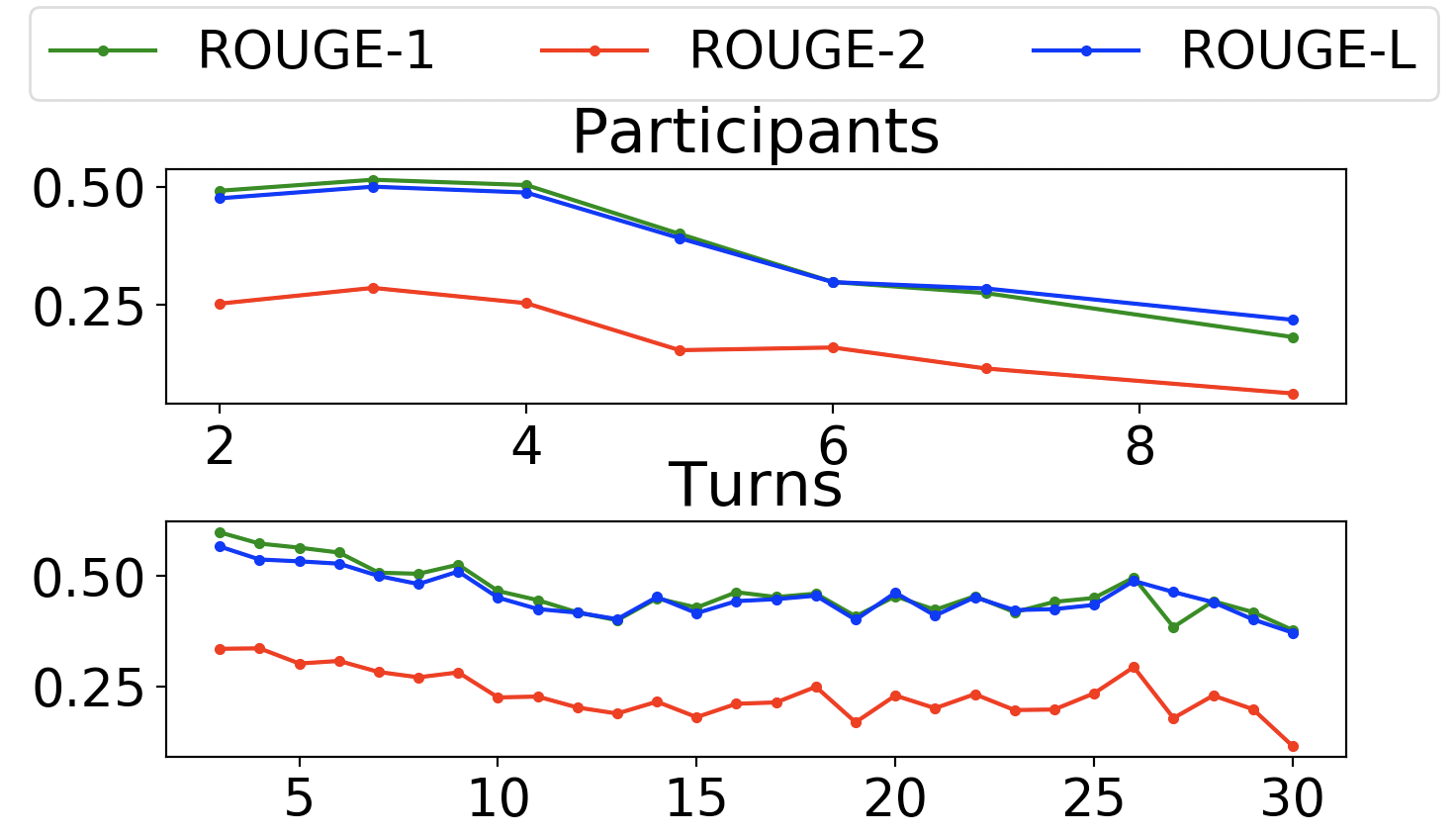}
\caption{Relations between  ROUGE scores and the number of participants/turns in conversations.}\label{Fig:changing_roles_rounds}
\end{figure}

We did not see any performance boost when combining the  generic global view with either topic or conversational stage views, partially due to that the coarse granularity of global view does not complement structured views well.  In contrast, utilizing both structured views (topic view + stage view) further increased ROUGE scores consistently, indicating the effectiveness of synthesizing informative conversation blocks introduced by both views.

We visualized the attention weight distributions for the stage view and topic view in our best model (see Appendix) and found  contributions of topic views are slightly more prominent compared to stage views. This also communicated that the two different structured views can complement each other well though sharing the same dialogue content. 
Note that the gains from \textsl{Multi-view BART (Topic + Stage)} are mainly from the precision scores while recall scores are kept comparable, suggesting that our proposed model produced fewer irrelevant tokens while preserving necessary information in its generated summary.

\paragraph{Impact of Participants and Turns}
We visualized the impact of two essential components in conversations---the number of participants and turns---on rouge scores via our best-performing model \textsl{Multi-view BART with topic view + stage view} in Figure~\ref{Fig:changing_roles_rounds}. As the number of participants/turns increases, ROUGE scores decrease, indicating that the difficulty of conversation summarization increased with more participants involved in conversations and more utterances.

\begin{figure}[t]
\centering
\includegraphics[width=0.95\columnwidth]{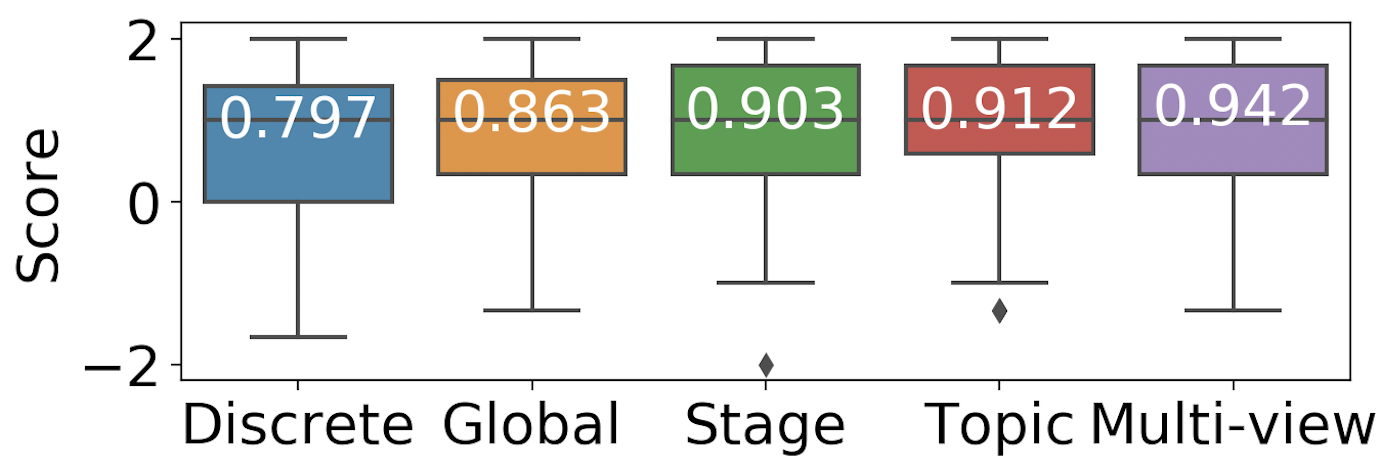}
\caption{Human evaluation results. The mean score for each model is also shown in the box plot. 
}\label{Fig:human_eval}
\end{figure}

\paragraph{Qualitative/Human Evaluation}
We also conducted human annotations to evaluate the extracted dialogue summaries, in addition to ROUGE scores. Similar to \citet{gliwa2019samsum}, we asked human annotators on Amazon Mechanical Turk \footnote{\url{https://www.mturk.com/}} to rate each summary (200 randomly sampled summaries in total) on the scale of [-2, 0, 2], where -2 means that a summary was poor, extracted irrelevant information or did not make sense at all, 2 means it was understandable and gave a concise overview of the text, and 0 refers to that the summary only extracted only a part of relevant information, or made some mistakes. The score for each summary was averaged among three different annotators. The Intra-class Correlation was 0.583, indicating moderate agreement \cite{koo2016guideline}. 

As shown in Figure~\ref{Fig:human_eval}, consistent with ROUGE scores in Table \ref{Tab: Main_results}, our multi-view model achieved the highest human annotation scores, significantly higher (via a student t-test) than either generic (discrete or global) view or structured (stage or topic) view, which further proved the effectiveness of combing different views.

\section{Model Analysis and Discussion}
So far, we have achieved a reasonable summarization performance.
To further study why dialog summarization is challenging and how future research could advance this direction, we take a closer look at this dialogue summarization dataset (SAMSum), model generation errors, as well as certain challenges that existing approaches are struggling with. 

\subsection{Challenges in Dialog Summarization}
We conduct a thorough examination of the challenges in conversation summarization and organized them into 7 categories as below: 

\begin{enumerate}
    \item \textbf{Informal language use} Many conversations especially in online contexts such as Twitter/Reddit \cite{jackson2007natural}, contain typos, word abbreviations, slang or emoticons/emojis, making it hard to be represented and summarized. 
    \item \textbf{Multiple participants} As shown in Figure~\ref{Fig:changing_roles_rounds}, conversations with more speakers are harder to be summarized since it may require models to accurately differentiate both language styles and content from different speakers, similar to the multiple characters issue in story summarization \cite{zhang2019generating}.
    \item \textbf{Multiple turns} Similar to long document summarization \cite{Xiao2019ExtractiveSO}, conversations with many utterances contain more information to be processed, thus harder to be summarized. 
    \item \textbf{(Referral and coreference}  People usually refer to each other, mention others' names or use coreference in their messages, which introduces extra difficulty to dialogue summarization, also a challenge also exists in reading comprehension \cite{chen-etal-2016-thorough} and document summarization \cite{falke2017concept}.
    \item \textbf{Repetition and interruption} Information is generally scattered through the whole conversation, and speakers may interrupt each other, reconfirm, back channeling or repeat themselves, a unique discourse challenge for dialogue summarization. 
    \item \textbf{Negations and rhetorical questions} As a long-standing problem in NLP field \cite{li2015visualizing}, negation related issues are even more frequent in conversations, as there are more question-answer exchanges between speakers. 
    \item \textbf{Role and language change} Conversations usually involve more than one speaker, and the role of a speaker may shift from a questioner to an answerer, requiring the summarization model to dynamically deal with speaker roles and the associated language (e.g., first personal pronouns)
\end{enumerate}

We randomly sampled 100 examples\footnote{The full analyzed set of examples are shown in Appendix.} from our test set and classified them using the above challenge taxonomy. A conversation might have more than one category labels, and if it had none of the aforementioned challenges, we labeled it as
\textbf{(0) Generic}. Usually, the one marked as \textsl{Generic} were shorter or had a simple structure.

Table~\ref{Tab:difficulty} presents the percentage of each type of challenge and per-category performances from our best model (\textsl{Multi-view BART with Topic view + Stage view}).
We observed that: 
(i) \textsl{Referral \& coreference} (33\%) and \textsl{Role \& language change} (30\%) were the two most frequent challenges that dialogue summarization task faced. 
(2) As expected, \textsl{Generic} conversations were relatively easier summarize.
(3) Our best model performed relatively worse when it came to \textsl{Repetition \& interruption}, \textsl{Multiple turns}, and \textsl{Referral \& coreference}, calling for more intelligent summarization methods to tackle those challenges.

\begin{table}[t]
\small
\centering
\begin{tabular}{|r|c|c|}
\hline
\textbf{Challenge} & {\textbf{\%}} & {\textbf{ ROUGE-1/2/L}}  \\ \hline
Generic                     & 24                                 & \textbf{0.613 / 0.384 / 0.579}                       \\ \hline
Informal language                       & 25                                 & 0.471 / 0.241 / 0.459                               \\
Multiple participants                    & 10                                 & 0.473 / 0.243 / 0.461                                \\
Multiple turns                   & 23                                 & 0.432 / 0.213 / 0.432                       \\
Referral \& coreference                    & 33                                 & 0.445 / 0.206  / 0.430                                 \\
Repetition \& interruption                   & 18                                 & 0.423 / 0.180 / 0.415                     \\
Negations \& rhetorical                   & 20                                 & 0.458 / 0.227 / 0.431                                \\
Role \& language change                   & 30                                 & 0.469 / 0.211 / 0.450     \\ \hline                            
\end{tabular} \caption{ The breakdown of challenges in dialogue summarization based on our analyses of 100 sampled conversations, and the ROUGE scores per challenge} \label{Tab:difficulty}
\end{table}
\subsection{Error Analysis\footnote{Error analysis for baselines are displayed in the Appendix.}}
We examined summaries generated by our best-performing model compared to ground-truth summaries, and observed several major error types:

\begin{table}[t]
\centering
\small
\begin{tabular}{|r|c|c|}
\hline
\textbf{Errors} & \multicolumn{1}{l|}{\textbf{\%}} & \multicolumn{1}{l|}{\textbf{ ROUGE-1/2/L}}  \\ \hline
Other             & 24                                  & \textbf{0.611 / 0.363 / 0.584}                         \\ \hline
Missing information             & 37                                  & 0.448 / 0.236 / 0.445                                 \\
Redundancy             & 13                                  & 0.442 / 0.231 / 0.441                               \\ 
Wrong references             & 27                                  & 0.460 / 0.232 / 0.454                             \\
Incorrect reasoning             & 24                                  & 0.447 /0.187 / 0.411                                 \\ 
Improper gendered pronouns             & 6                                   & 0.421 / 0.212 / 0.428                                  \\ \hline
\end{tabular} \caption{The common error types of our model compared to golden reference on 100 sampled conversations, and the ROUGE scores per error type.} \label{Tab:errors}
\end{table}

\begin{enumerate}
    \item \textbf{Missing information}: content mentioned in references is missing in generated summaries.
    \item \textbf{Redundancy}:  content occurred in generated summaries was not mentioned by references.
    \item \textbf{Wrong references}: generated summaries contain information that is not faithful to the original dialogue, and associate one's actions/locations with a wrong speaker.
    \item \textbf{Incorrect reasoning}: generated summaries reasoned relations in dialogues incorrectly, thus came to wrong conclusions.
    \item \textbf{Improper gendered pronouns}: summaries used improper gendered pronouns (e.g., the misuse of gendered pronouns).
\end{enumerate}

We annotated the same set of 100 randomly sampled summaries via the above error type taxonomy.  A summary might have more than one category labels and we categorized a summary as \textbf{(0) Other} if it did not belong to any error types.
 
Table~\ref{Tab:errors} presents the breakdown of error types and per-category ROUGE scores.
We found that: (i) \textsl{missing information} (37\%) was the most frequent error type, indicating that current summarization models struggled with identifying key information. 
(ii) \textsl{Incorrect reasoning} had a percentage of 24\% with the worst ROUGE-2; despite of being a minor type 6\%, \textsl{improper gendered pronouns} seemed to severely decrease both ROUGE-1 and ROUGE-2. 
(iii) The relatively low ROUGE scores associated with \textsl{incorrect reasoning} and \textsl{wrong references} urged better summarization models in dealing with faithfulness in dialogue summarization.  

\subsection{Relation between Challenges and Errors}
\begin{figure}[t]
\centering
\includegraphics[width=0.7\columnwidth]{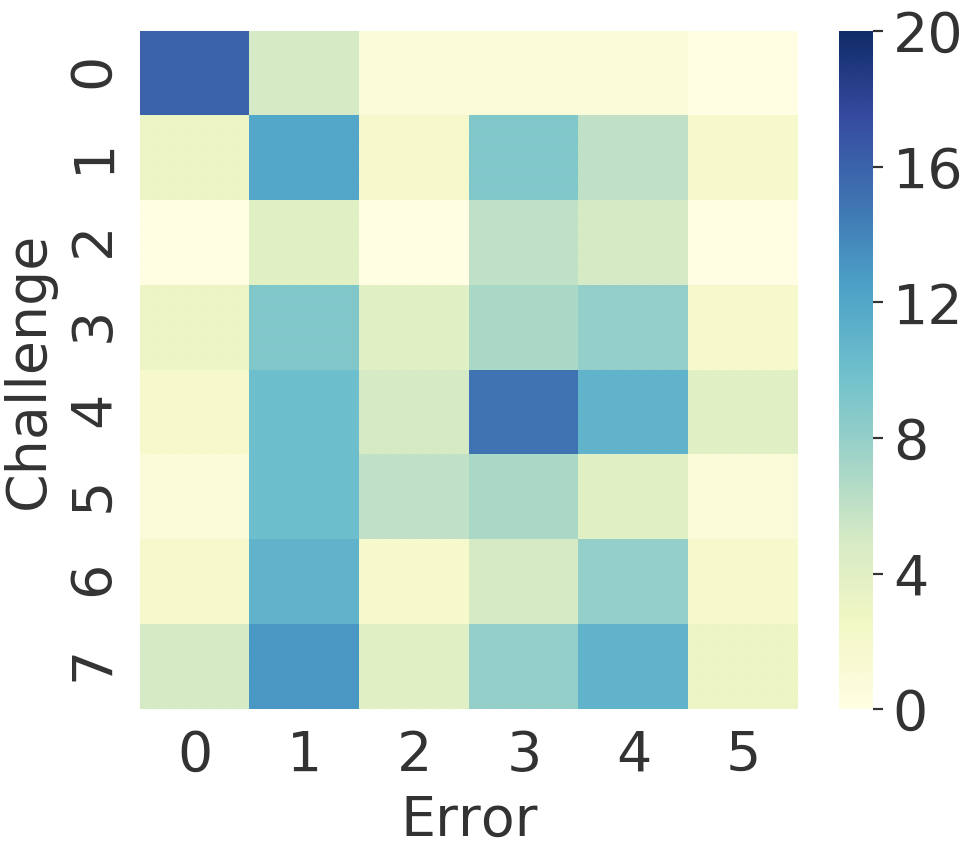}
\caption{Relations between difficulties in conversations and errors made by our model.}\label{Fig:error_relations}
\end{figure}

To figure out relations between challenges and errors made by our models, i.e., how different types of errors correlate with different types of challenges, we visualized the co-occurrence heat map in Figure~\ref{Fig:error_relations}. 
We found that: (i) Our model generated good summary for \textsl{generic}, simple conversations. (ii) All kinds of challenges had high correlations with, or could lead to the \textsl{missing information} error. (iii) \textsl{Wrong references} were highly associated with \textsl{referral \& coreference}; this was as expected since co-references in conversations would naturally increase the difficulty for models to associate correct speakers with correct actions.
(iv) High correlations between \textsl{role \& language change}, \textsl{referral \& coreference} and \textsl{incorrect reasoning} indicated that interactions between multiple participants with frequent co-references might easily lead current summarization models to reason incorrectly.
\section{Conclusion}
In this work, we proposed a multi-view sequence-to-sequence model that leveraged multiple conversational structures (topic view and stage view) and  generic views (global view and discrete view) to generate summaries for conversations. 
In order to strategically combine these different views for better summary generations, we propose a multi-view sequence-to-sequence model. 
Experiments conducted demonstrated the effectiveness of our proposed models in terms of both quantitative and qualitative evaluations. Via thorough error analyses, we concluded a set of challenges that current models struggled with, which can further facilitate future research on conversation summarization. Due to the lack of annotations, we only adopted simple unsupervised segmentation methods to extract different views. In the future, we plan to annotate some of the data,  explore supervised segmentation models \cite{li2018segbot} and introduce more conversation structures like dialogue acts \cite{oya2014extractive,joty2016speech} into abstractive dialogue summarization.

\section*{Acknowledgment }
We would like to thank the anonymous reviewers
for their helpful comments, and the members of Georgia Tech SALT group for their feedback. We acknowledge the support of NVIDIA Corporation with the donation of GPU used for this research. 

\bibliography{emnlp2020}

\begin{thebibliography}{51}
\expandafter\ifx\csname natexlab\endcsname\relax\def\natexlab#1{#1}\fi

\bibitem[{Althoff et~al.(2016)Althoff, Clark, and
  Leskovec}]{althoff-etal-2016-large}
Tim Althoff, Kevin Clark, and Jure Leskovec. 2016.
\newblock \href {https://doi.org/10.1162/tacl_a_00111} {Large-scale analysis of
  counseling conversations: An application of natural language processing to
  mental health}.
\newblock \emph{Transactions of the Association for Computational Linguistics},
  4:463--476.

\bibitem[{Beltagy et~al.(2020)Beltagy, Peters, and
  Cohan}]{beltagy2020longformer}
Iz~Beltagy, Matthew~E. Peters, and Arman Cohan. 2020.
\newblock \href {http://arxiv.org/abs/2004.05150} {Longformer: The
  long-document transformer}.

\bibitem[{Cao et~al.(2018)Cao, Wei, Li, and Li}]{Cao2018FaithfulTT}
Ziqiang Cao, Furu Wei, Wenjie Li, and Sujian Li. 2018.
\newblock Faithful to the original: Fact aware neural abstractive
  summarization.
\newblock In \emph{AAAI}.

\bibitem[{Chen et~al.(2016)Chen, Bolton, and Manning}]{chen-etal-2016-thorough}
Danqi Chen, Jason Bolton, and Christopher~D. Manning. 2016.
\newblock \href {https://doi.org/10.18653/v1/P16-1223} {A thorough examination
  of the {CNN}/daily mail reading comprehension task}.
\newblock In \emph{Proceedings of the 54th Annual Meeting of the Association
  for Computational Linguistics (Volume 1: Long Papers)}, pages 2358--2367,
  Berlin, Germany. Association for Computational Linguistics.

\bibitem[{Chen and Bansal(2018)}]{DBLP:journals/corr/abs-1805-11080}
Yen-Chun Chen and Mohit Bansal. 2018.
\newblock \href {https://doi.org/10.18653/v1/P18-1063} {Fast abstractive
  summarization with reinforce-selected sentence rewriting}.
\newblock In \emph{Proceedings of the 56th Annual Meeting of the Association
  for Computational Linguistics (Volume 1: Long Papers)}, pages 675--686,
  Melbourne, Australia. Association for Computational Linguistics.

\bibitem[{Choi(2000)}]{choi-2000-advances}
Freddy Y.~Y. Choi. 2000.
\newblock \href {https://www.aclweb.org/anthology/A00-2004} {Advances in domain
  independent linear text segmentation}.
\newblock In \emph{1st Meeting of the North {A}merican Chapter of the
  Association for Computational Linguistics}.

\bibitem[{Cohan et~al.(2018)Cohan, Dernoncourt, Kim, Bui, Kim, Chang, and
  Goharian}]{cohan-etal-2018-discourse}
Arman Cohan, Franck Dernoncourt, Doo~Soon Kim, Trung Bui, Seokhwan Kim, Walter
  Chang, and Nazli Goharian. 2018.
\newblock \href {https://doi.org/10.18653/v1/N18-2097} {A discourse-aware
  attention model for abstractive summarization of long documents}.
\newblock In \emph{Proceedings of the 2018 Conference of the North {A}merican
  Chapter of the Association for Computational Linguistics: Human Language
  Technologies, Volume 2 (Short Papers)}, pages 615--621, New Orleans,
  Louisiana. Association for Computational Linguistics.

\bibitem[{Fabbri et~al.(2019)Fabbri, Li, She, Li, and Radev}]{Fabbri_2019}
Alexander Fabbri, Irene Li, Tianwei She, Suyi Li, and Dragomir Radev. 2019.
\newblock \href {https://doi.org/10.18653/v1/p19-1102} {Multi-news: A
  large-scale multi-document summarization dataset and abstractive hierarchical
  model}.
\newblock \emph{Proceedings of the 57th Annual Meeting of the Association for
  Computational Linguistics}.

\bibitem[{Falke et~al.(2017)Falke, Meyer, and Gurevych}]{falke2017concept}
Tobias Falke, Christian~M Meyer, and Iryna Gurevych. 2017.
\newblock Concept-map-based multi-document summarization using concept
  coreference resolution and global importance optimization.
\newblock In \emph{Proceedings of the Eighth International Joint Conference on
  Natural Language Processing (Volume 1: Long Papers)}, pages 801--811.

\bibitem[{Galley et~al.(2003)Galley, McKeown, Fosler-Lussier, and
  Jing}]{galley-etal-2003-discourse}
Michel Galley, Kathleen~R. McKeown, Eric Fosler-Lussier, and Hongyan Jing.
  2003.
\newblock \href {https://doi.org/10.3115/1075096.1075167} {Discourse
  segmentation of multi-party conversation}.
\newblock In \emph{Proceedings of the 41st Annual Meeting of the Association
  for Computational Linguistics}, pages 562--569, Sapporo, Japan. Association
  for Computational Linguistics.

\bibitem[{Gao et~al.(2020)Gao, Chen, Ren, Zhao, and Yan}]{gao2020standard}
Shen Gao, Xiuying Chen, Zhaochun Ren, Dongyan Zhao, and Rui Yan. 2020.
\newblock \href {http://arxiv.org/abs/2005.04684} {From standard summarization
  to new tasks and beyond: Summarization with manifold information}.

\bibitem[{Gliwa et~al.(2019)Gliwa, Mochol, Biesek, and Wawer}]{gliwa2019samsum}
Bogdan Gliwa, Iwona Mochol, Maciej Biesek, and Aleksander Wawer. 2019.
\newblock \href {https://doi.org/10.18653/v1/D19-5409} {{SAMS}um corpus: A
  human-annotated dialogue dataset for abstractive summarization}.
\newblock In \emph{Proceedings of the 2nd Workshop on New Frontiers in
  Summarization}, pages 70--79, Hong Kong, China. Association for Computational
  Linguistics.

\bibitem[{Goo and Chen(2018)}]{Goo_2018}
Chih-Wen Goo and Yun-Nung Chen. 2018.
\newblock \href {https://doi.org/10.1109/slt.2018.8639531} {Abstractive
  dialogue summarization with sentence-gated modeling optimized by dialogue
  acts}.
\newblock \emph{2018 IEEE Spoken Language Technology Workshop (SLT)}.

\bibitem[{Hochreiter and Schmidhuber(1997)}]{10.1162/neco.1997.9.8.1735}
Sepp Hochreiter and J\"{u}rgen Schmidhuber. 1997.
\newblock \href {https://doi.org/10.1162/neco.1997.9.8.1735} {Long short-term
  memory}.
\newblock \emph{Neural Comput.}, 9(8):1735–1780.

\bibitem[{Honneth et~al.(1988)Honneth, Joas et~al.}]{honneth1988social}
Axel Honneth, Hans Joas, et~al. 1988.
\newblock \emph{Social action and human nature}.
\newblock CUP Archive.

\bibitem[{Jackson and Moulinier(2007)}]{jackson2007natural}
Peter Jackson and Isabelle Moulinier. 2007.
\newblock \emph{Natural language processing for online applications: Text
  retrieval, extraction and categorization}, volume~5.
\newblock John Benjamins Publishing.

\bibitem[{Joty et~al.(2010)Joty, Carenini, Murray, and
  Ng}]{joty-etal-2010-exploiting}
Shafiq Joty, Giuseppe Carenini, Gabriel Murray, and Raymond~T. Ng. 2010.
\newblock \href {https://www.aclweb.org/anthology/D10-1038} {Exploiting
  conversation structure in unsupervised topic segmentation for emails}.
\newblock In \emph{Proceedings of the 2010 Conference on Empirical Methods in
  Natural Language Processing}, pages 388--398, Cambridge, MA. Association for
  Computational Linguistics.

\bibitem[{Joty and Hoque(2016)}]{joty2016speech}
Shafiq Joty and Enamul Hoque. 2016.
\newblock Speech act modeling of written asynchronous conversations with
  task-specific embeddings and conditional structured models.
\newblock In \emph{Proceedings of the 54th Annual Meeting of the Association
  for Computational Linguistics (Volume 1: Long Papers)}, pages 1746--1756.

\bibitem[{Kester(2004)}]{kester2004conversation}
Grant~H Kester. 2004.
\newblock \emph{Conversation pieces: Community and communication in modern
  art}.
\newblock Univ of California Press.

\bibitem[{Koo and Li(2016)}]{koo2016guideline}
Terry~K Koo and Mae~Y Li. 2016.
\newblock A guideline of selecting and reporting intraclass correlation
  coefficients for reliability research.
\newblock \emph{Journal of chiropractic medicine}, 15(2):155--163.

\bibitem[{Lewis et~al.(2019)Lewis, Liu, Goyal, Ghazvininejad, Mohamed, Levy,
  Stoyanov, and Zettlemoyer}]{lewis2019bart}
Mike Lewis, Yinhan Liu, Naman Goyal, Marjan Ghazvininejad, Abdelrahman Mohamed,
  Omer Levy, Ves Stoyanov, and Luke Zettlemoyer. 2019.
\newblock \href {http://arxiv.org/abs/1910.13461} {Bart: Denoising
  sequence-to-sequence pre-training for natural language generation,
  translation, and comprehension}.

\bibitem[{Li et~al.(2018)Li, Sun, and Joty}]{li2018segbot}
Jing Li, Aixin Sun, and Shafiq~R Joty. 2018.
\newblock Segbot: A generic neural text segmentation model with pointer
  network.
\newblock In \emph{IJCAI}, pages 4166--4172.

\bibitem[{Li et~al.(2016)Li, Chen, Hovy, and Jurafsky}]{li2015visualizing}
Jiwei Li, Xinlei Chen, Eduard Hovy, and Dan Jurafsky. 2016.
\newblock \href {https://doi.org/10.18653/v1/N16-1082} {Visualizing and
  understanding neural models in {NLP}}.
\newblock In \emph{Proceedings of the 2016 Conference of the North {A}merican
  Chapter of the Association for Computational Linguistics: Human Language
  Technologies}, pages 681--691, San Diego, California. Association for
  Computational Linguistics.

\bibitem[{Li et~al.(2019)Li, Zhang, Ji, and Radke}]{li-etal-2019-keep}
Manling Li, Lingyu Zhang, Heng Ji, and Richard~J. Radke. 2019.
\newblock \href {https://doi.org/10.18653/v1/P19-1210} {Keep meeting summaries
  on topic: Abstractive multi-modal meeting summarization}.
\newblock In \emph{Proceedings of the 57th Annual Meeting of the Association
  for Computational Linguistics}, pages 2190--2196, Florence, Italy.
  Association for Computational Linguistics.

\bibitem[{Lin and Och(2004)}]{lin2004automatic}
Chin-Yew Lin and Franz~Josef Och. 2004.
\newblock Automatic evaluation of machine translation quality using longest
  common subsequence and skip-bigram statistics.
\newblock In \emph{Proceedings of the 42nd Annual Meeting on Association for
  Computational Linguistics}, page 605. Association for Computational
  Linguistics.

\bibitem[{Liu et~al.(2019{\natexlab{a}})Liu, Wang, Xu, Li, and
  Ye}]{10.1145/3292500.3330683}
Chunyi Liu, Peng Wang, Jiang Xu, Zang Li, and Jieping Ye. 2019{\natexlab{a}}.
\newblock \href {https://doi.org/10.1145/3292500.3330683} {Automatic dialogue
  summary generation for customer service}.
\newblock In \emph{Proceedings of the 25th ACM SIGKDD International Conference
  on Knowledge Discovery \& Data Mining}, KDD19, page 1957–1965, New York,
  NY, USA. Association for Computing Machinery.

\bibitem[{Liu* et~al.(2018)Liu*, Saleh*, Pot, Goodrich, Sepassi, Kaiser, and
  Shazeer}]{DBLP:journals/corr/abs-1801-10198}
Peter~J. Liu*, Mohammad Saleh*, Etienne Pot, Ben Goodrich, Ryan Sepassi, Lukasz
  Kaiser, and Noam Shazeer. 2018.
\newblock \href {https://openreview.net/forum?id=Hyg0vbWC-} {Generating
  wikipedia by summarizing long sequences}.
\newblock In \emph{International Conference on Learning Representations}.

\bibitem[{Liu and Lapata(2019)}]{Liu_2019_pretrained}
Yang Liu and Mirella Lapata. 2019.
\newblock \href {https://doi.org/10.18653/v1/d19-1387} {Text summarization with
  pretrained encoders}.
\newblock \emph{Proceedings of the 2019 Conference on Empirical Methods in
  Natural Language Processing and the 9th International Joint Conference on
  Natural Language Processing (EMNLP-IJCNLP)}.

\bibitem[{Liu and Chen(2019)}]{liu-chen-2019-reading}
Zhengyuan Liu and Nancy Chen. 2019.
\newblock \href {https://doi.org/10.18653/v1/P19-1543} {Reading turn by turn:
  Hierarchical attention architecture for spoken dialogue comprehension}.
\newblock In \emph{Proceedings of the 57th Annual Meeting of the Association
  for Computational Linguistics}, pages 5460--5466, Florence, Italy.
  Association for Computational Linguistics.

\bibitem[{Liu et~al.(2019{\natexlab{b}})Liu, Ng, Lee, Aw, and Chen}]{Liu_2019}
Zhengyuan Liu, Angela Ng, Sheldon Lee, Ai~Ti Aw, and Nancy~F. Chen.
  2019{\natexlab{b}}.
\newblock \href {https://doi.org/10.1109/asru46091.2019.9003764} {Topic-aware
  pointer-generator networks for summarizing spoken conversations}.
\newblock \emph{2019 IEEE Automatic Speech Recognition and Understanding
  Workshop (ASRU)}.

\bibitem[{Mayfield et~al.(2012)Mayfield, Adamson, and
  Penstein~Ros{\'e}}]{mayfield-etal-2012-hierarchical}
Elijah Mayfield, David Adamson, and Carolyn Penstein~Ros{\'e}. 2012.
\newblock \href {https://www.aclweb.org/anthology/W12-1607} {Hierarchical
  conversation structure prediction in multi-party chat}.
\newblock In \emph{Proceedings of the 13th Annual Meeting of the Special
  Interest Group on Discourse and Dialogue}, pages 60--69, Seoul, South Korea.
  Association for Computational Linguistics.

\bibitem[{Nallapati et~al.(2016)Nallapati, Zhou, dos Santos,
  GuÌ‡l{\c{c}}ehre, and Xiang}]{nallapati-etal-2016-abstractive}
Ramesh Nallapati, Bowen Zhou, Cicero dos Santos, {\c{C}}a{\u{g}}lar
  GuÌ‡l{\c{c}}ehre, and Bing Xiang. 2016.
\newblock \href {https://doi.org/10.18653/v1/K16-1028} {Abstractive text
  summarization using sequence-to-sequence {RNN}s and beyond}.
\newblock In \emph{Proceedings of The 20th {SIGNLL} Conference on Computational
  Natural Language Learning}, pages 280--290, Berlin, Germany. Association for
  Computational Linguistics.

\bibitem[{Nikolov et~al.(2018)Nikolov, Pfeiffer, and
  Hahnloser}]{DBLP:journals/corr/abs-1804-08875}
Nikola~I. Nikolov, Michael Pfeiffer, and Richard H.~R. Hahnloser. 2018.
\newblock \href {http://arxiv.org/abs/1804.08875} {Data-driven summarization of
  scientific articles}.
\newblock \emph{CoRR}, abs/1804.08875.

\bibitem[{Oya and Carenini(2014)}]{oya2014extractive}
Tatsuro Oya and Giuseppe Carenini. 2014.
\newblock Extractive summarization and dialogue act modeling on email threads:
  An integrated probabilistic approach.
\newblock In \emph{Proceedings of the 15th Annual Meeting of the Special
  Interest Group on Discourse and Dialogue (SIGDIAL)}, pages 133--140.

\bibitem[{Paul(2012)}]{paul-2012-mixed}
Michael~J. Paul. 2012.
\newblock \href {https://www.aclweb.org/anthology/D12-1009} {Mixed membership
  {M}arkov models for unsupervised conversation modeling}.
\newblock In \emph{Proceedings of the 2012 Joint Conference on Empirical
  Methods in Natural Language Processing and Computational Natural Language
  Learning}, pages 94--104, Jeju Island, Korea. Association for Computational
  Linguistics.

\bibitem[{Paulus et~al.(2018)Paulus, Xiong, and Socher}]{paulus2018a}
Romain Paulus, Caiming Xiong, and Richard Socher. 2018.
\newblock \href {https://openreview.net/forum?id=HkAClQgA-} {A deep reinforced
  model for abstractive summarization}.
\newblock In \emph{International Conference on Learning Representations}.

\bibitem[{Radford et~al.(2019)Radford, Wu, Child, Luan, Amodei, and
  Sutskever}]{radford2019language}
Alec Radford, Jeffrey Wu, Rewon Child, David Luan, Dario Amodei, and Ilya
  Sutskever. 2019.
\newblock Language models are unsupervised multitask learners.
\newblock \emph{OpenAI Blog}, 1(8):9.

\bibitem[{Raffel et~al.(2019)Raffel, Shazeer, Roberts, Lee, Narang, Matena,
  Zhou, Li, and Liu}]{raffel2019exploring}
Colin Raffel, Noam Shazeer, Adam Roberts, Katherine Lee, Sharan Narang, Michael
  Matena, Yanqi Zhou, Wei Li, and Peter~J. Liu. 2019.
\newblock \href {http://arxiv.org/abs/1910.10683} {Exploring the limits of
  transfer learning with a unified text-to-text transformer}.

\bibitem[{Reimers and Gurevych(2019)}]{Reimers_2019}
Nils Reimers and Iryna Gurevych. 2019.
\newblock \href {https://doi.org/10.18653/v1/d19-1410} {Sentence-bert: Sentence
  embeddings using siamese bert-networks}.
\newblock \emph{Proceedings of the 2019 Conference on Empirical Methods in
  Natural Language Processing and the 9th International Joint Conference on
  Natural Language Processing (EMNLP-IJCNLP)}.

\bibitem[{Ritter et~al.(2010)Ritter, Cherry, and
  Dolan}]{ritter-etal-2010-unsupervised}
Alan Ritter, Colin Cherry, and Bill Dolan. 2010.
\newblock \href {https://www.aclweb.org/anthology/N10-1020} {Unsupervised
  modeling of twitter conversations}.
\newblock In \emph{Human Language Technologies: The 2010 Annual Conference of
  the North {A}merican Chapter of the Association for Computational
  Linguistics}, pages 172--180, Los Angeles, California. Association for
  Computational Linguistics.

\bibitem[{Rush et~al.(2015)Rush, Chopra, and Weston}]{rush-etal-2015-neural}
Alexander~M. Rush, Sumit Chopra, and Jason Weston. 2015.
\newblock \href {https://doi.org/10.18653/v1/D15-1044} {A neural attention
  model for abstractive sentence summarization}.
\newblock In \emph{Proceedings of the 2015 Conference on Empirical Methods in
  Natural Language Processing}, pages 379--389, Lisbon, Portugal. Association
  for Computational Linguistics.

\bibitem[{Sacks et~al.(1978)Sacks, Schegloff, and
  Jefferson}]{sacks1978simplest}
Harvey Sacks, Emanuel~A Schegloff, and Gail Jefferson. 1978.
\newblock A simplest systematics for the organization of turn taking for
  conversation.
\newblock In \emph{Studies in the organization of conversational interaction},
  pages 7--55. Elsevier.

\bibitem[{See et~al.(2017)See, Liu, and Manning}]{See_2017}
Abigail See, Peter~J. Liu, and Christopher~D. Manning. 2017.
\newblock \href {https://doi.org/10.18653/v1/p17-1099} {Get to the point:
  Summarization with pointer-generator networks}.
\newblock \emph{Proceedings of the 55th Annual Meeting of the Association for
  Computational Linguistics (Volume 1: Long Papers)}.

\bibitem[{Shang et~al.(2018)Shang, Ding, Zhang, Tixier, Meladianos,
  Vazirgiannis, and Lorr{\'e}}]{shang-etal-2018-unsupervised}
Guokan Shang, Wensi Ding, Zekun Zhang, Antoine Tixier, Polykarpos Meladianos,
  Michalis Vazirgiannis, and Jean-Pierre Lorr{\'e}. 2018.
\newblock \href {https://doi.org/10.18653/v1/P18-1062} {Unsupervised
  abstractive meeting summarization with multi-sentence compression and
  budgeted submodular maximization}.
\newblock In \emph{Proceedings of the 56th Annual Meeting of the Association
  for Computational Linguistics (Volume 1: Long Papers)}, pages 664--674,
  Melbourne, Australia. Association for Computational Linguistics.

\bibitem[{Wu et~al.(2019)Wu, Fan, Baevski, Dauphin, and
  Auli}]{DBLP:journals/corr/abs-1901-10430}
Felix Wu, Angela Fan, Alexei Baevski, Yann Dauphin, and Michael Auli. 2019.
\newblock \href {https://openreview.net/forum?id=SkVhlh09tX} {Pay less
  attention with lightweight and dynamic convolutions}.
\newblock In \emph{International Conference on Learning Representations}.

\bibitem[{Xiao and Carenini(2019)}]{Xiao2019ExtractiveSO}
Wen Xiao and Giuseppe Carenini. 2019.
\newblock Extractive summarization of long documents by combining global and
  local context.
\newblock In \emph{EMNLP/IJCNLP}.

\bibitem[{Zhang et~al.(2019)Zhang, Cheung, and Oren}]{zhang2019generating}
Weiwei Zhang, Jackie Chi~Kit Cheung, and Joel Oren. 2019.
\newblock Generating character descriptions for automatic summarization of
  fiction.
\newblock In \emph{Proceedings of the AAAI Conference on Artificial
  Intelligence}, volume~33, pages 7476--7483.

\bibitem[{Zhao et~al.(2019)Zhao, Pan, Fan, Liu, Li, Yang, and
  Cai}]{10.1145/3308558.3313619}
Zhou Zhao, Haojie Pan, Changjie Fan, Yan Liu, Linlin Li, Min Yang, and Deng
  Cai. 2019.
\newblock \href {https://doi.org/10.1145/3308558.3313619} {Abstractive meeting
  summarization via hierarchical adaptive segmental network learning}.
\newblock In \emph{The World Wide Web Conference}, WWW ’19, page 3455–3461,
  New York, NY, USA. Association for Computing Machinery.

\bibitem[{Zhu et~al.(2020{\natexlab{a}})Zhu, Hinthorn, Xu, Zeng, Zeng, Huang,
  and Jiang}]{zhu2020boosting}
Chenguang Zhu, William Hinthorn, Ruochen Xu, Qingkai Zeng, Michael Zeng,
  Xuedong Huang, and Meng Jiang. 2020{\natexlab{a}}.
\newblock \href {http://arxiv.org/abs/2003.08612} {Boosting factual correctness
  of abstractive summarization}.

\bibitem[{Zhu et~al.(2020{\natexlab{b}})Zhu, Xu, Zeng, and
  Huang}]{zhu2020endtoend}
Chenguang Zhu, Ruochen Xu, Michael Zeng, and Xuedong Huang. 2020{\natexlab{b}}.
\newblock \href {http://arxiv.org/abs/2004.02016} {End-to-end abstractive
  summarization for meetings}.

\bibitem[{Zhu et~al.(2019)Zhu, Nan, Wang, Nallapati, and Xiang}]{zhu2019did}
Henghui Zhu, Feng Nan, Zhiguo Wang, Ramesh Nallapati, and Bing Xiang. 2019.
\newblock \href {http://arxiv.org/abs/1911.10666} {Who did they respond to?
  conversation structure modeling using masked hierarchical transformer}.

\end{thebibliography}
\bibliographystyle{acl_natbib}

\appendix

\section{Model Settings}
We load the pre-trained ``bert-base-nli-stsb-mean-tokens"\footnote{\url{https://github.com/UKPLab/sentence-transformers}} for sentence-BERT to get representations for each utterance. When extracting the \textbf{topic view}, we set the window size 4 and std coefficient 1 in C99. When extracting the \textbf{stage view}, we set the number of hidden states 4 in HMM. These hyper-parameters were set after a grid search with evaluating randomly sampled segmented results by human.  The \textbf{BART + Structured views} (stage and topic views) followed the same parameters as \textbf{BART + Generic views}. 

\noindent
For \textbf{Multi-View BART}, we selected different views to combine: (1) generic view + structured view:  best generic view, global view, was combined with two structured views (stage and topic view); (2) structured view + structured view: best two single views are combined (topic + stage). The settings for BART encoder/decoder kept the same as baseline. We used a one layer LSTM for encoding sections. The learning rate for section encoder and multi-view attention was set 3e-3. The temperature $T$ was 0.2. The beam search size during inference for all the models was 4.

Experiments were performed on two Tesla P100 (16GB memory).

\section{View Attention Visualization}
\begin{figure}[t]
\centering
\includegraphics[width=0.8\columnwidth]{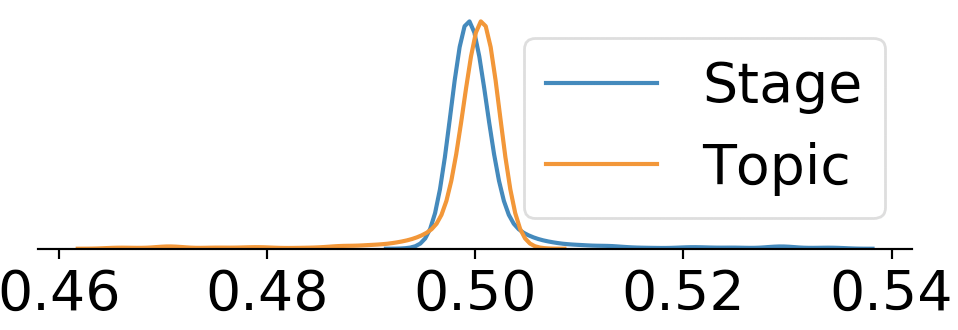}
\caption{Attention weights distribution for stage view and topic view in the multi-view model. }\label{Fig:attn_weights}
\end{figure}

We visualized the attention weights distribution for the stage view and topic view in our best multi-view model to explore the importance of stage verses topic in Figure~\ref{Fig:attn_weights}.We found that the topic views were more prominent than the stage views, consistent with the performances of \textsl{BART + topic view} and \textsl{BART + stage view}. This indicated that having discourse structures about topics might be more important while both topic and stage could improve the conversation summarization. This also communicated that the two different structured views can complement each other well though sharing the same dialogue content.

\begin{table}[t]
\centering
\begin{tabular}{|c|c|c|c|c|}
\hline
330                    & 191                    & 635                    & 733                    & 342                    \\
595                    & 454                    & 629                    & 598                    & 466                    \\
158                    & 576                    & 676                    & 344                    & 353                    \\
621                    & 255                    & 106                    & 66                     & 742                    \\
446                    & 327                    & 497                    & 463                    & 478                    \\
320                    & 258                    & 528                    & 405                    & 305                    \\
208                    & 550                    & 512                    & 663                    & 165                    \\
69                     & 431                    & 796                    & 338                    & 443                    \\
254                    & 716                    & 549                    & 51                     & 145                    \\
364                    & 259                    & 190                    & 479                    & 182                    \\
617                    & 189                    & 422                    & 177                    & 8                      \\
741                    & 151                    & 488                    & 176                    & 212                    \\
15                     & 124                    & 461                    & 386                    & 197                    \\
172                    & 372                    & 508                    & 323                    & 162                    \\
793                    & 308                    & 486                    & 763                    & 376                    \\
493                    & 520                    & 116                    & 513                    & 802                    \\
358                    & 784                    & 53                     & 655                    & 23                     \\
717                    & 374                    & 289                    & 64                     & 217                    \\
519                    & 539                    & 441                    & 341                    & 350                    \\
136                    & 713                    & 426                    & 648                    & 355        \\ \hline            
\end{tabular} \caption{A full index list of our samples.} \label{Tab:Full_idx}
\end{table}

We displayed two examples in Table~\ref{Tab:view_examples} with the golden references, each single view's generated summaries and the combined views' generated summaries. The combined view could balance the advantages of each single view and generated more precise summaries. And the attention weights the model learned were also consistent with single view's performances.

\begin{table*}[t]
\centering
\begin{tabular}{|c|l||l|}
\hline
\textbf{Reference}  & \begin{tabular}[c]{@{}l@{}}James misses Hannah. They agree for James \\to pick Hannah up on \\ Saturday at 8.\end{tabular}                                   & \begin{tabular}[c]{@{}l@{}}Petra is very sleepy at work today, \\Andy finds the day boring, \\ and Ezgi is working.\end{tabular}                              \\ \hline \hline
Stage         & \begin{tabular}[c]{@{}l@{}}Hannah has to get up early for wo-\\ -rk tomorrow. James will pick her \\ up at 8 on Saturday.\\ \color[HTML]{A61C00}{{[}0.61/0.13/0.40{]}}\end{tabular} & \begin{tabular}[c]{@{}l@{}}Petra needs to sleep, because \\ she's sleepy. Ezgi is working.\\ \color[HTML]{A61C00}{{[}0.37/0.16/0.38{]}}\end{tabular}                                   \\ \hline
Topic         & \begin{tabular}[c]{@{}l@{}}James and Hannah will see each \\ other on Saturday at 8.\\ \color[HTML]{A61C00}{{[}0.46/0.25/0.50{]}}\end{tabular}                                      & \begin{tabular}[c]{@{}l@{}}Nobody is working at the office \\ today. Ezgi is working. Petra is \\ sleepy and wants to sleep.\\ \color[HTML]{A61C00}{{[}0.53/0.19/0.53{]}}\end{tabular} \\ \hline
Stage + Topic & \begin{tabular}[c]{@{}l@{}}James will pick Hannah up on \\ Saturday at 8 pm.\\ \color[HTML]{A61C00}{{[}0.64/0.52/0.69{]}}\end{tabular}                                              & \begin{tabular}[c]{@{}l@{}}Petra is sleepy and needs to sleep. \\ Ezgi is working at the office.\\ \color[HTML]{A61C00}{{[}0.60/0.21/0.43{]}}\end{tabular}                             \\ \hline
Attention Weight   & \color[HTML]{3399FF}{{[}0.52, 0.48{]} }                                                                                                                                             & \color[HTML]{3399FF}{{[}0.45. 0.55{]} }     \\ \hline                                                                                                                                           
\end{tabular} \caption{Some generated summary examples compared to references. \textcolor[HTML]{A61C00}{[Rouge-1/Rouge-2/Rouge-L]} is shown after each summary, and \textcolor[HTML]{3399FF}{[stage weight/topic weight]} is displayed in the last row.} \label{Tab:view_examples}
\end{table*}

\begin{table*}[t]
\centering
\begin{tabular}{|r|c|c|c|c|c|c|}
\hline
\textbf{Errors} & \multicolumn{1}{l|}{\textbf{Discrete}}& \multicolumn{1}{l|}{\textbf{Global}} & \multicolumn{1}{l|}{\textbf{Stage}} & \multicolumn{1}{l|}{\textbf{Topic}} & \multicolumn{1}{l|}{\textbf{Multi-view}}  \\ \hline
Other            &16 &19 &21 &22 & 24                                                        \\ \hline
Missing information   &40  &46 &45  &42      & 37                                                             \\
Redundancy          &33 &44   &18 &25 & 13                                                             \\ 
Wrong references     &32  &33  &26  &30  & 27                                                              \\
Incorrect reasoning   &27     &28 &22 &28   & 24                                                                 \\ 
Improper gendered pronouns    &5   &6 &6 &6     & 6                                                                     \\ \hline
\end{tabular} \caption{Common error types of different models compared to golden reference on 100 sampled conversations.} \label{Tab:errors_all}
\end{table*}
\section{Supplementary Examples for Model Analysis and Discussion}
For the analysis in the \textbf{Model Analysis and Discussion} section in our paper, we randomly sampled 100 examples from the test set of the SAMSum dataset which can be downloaded here \footnote{\url{https://arxiv.org/abs/1911.12237}}. Table~\ref{Tab:Full_idx} provides a full index list of the samples.

Table~\ref{Tab:errors_all} shows the error analysis for \textit{BART-Discrete, BART-Global, BART-Stage, BART-Topic} and \textit{BART-Multi-view} models. It can be observed that, (i) without any explicit structures, discrete-view and global-view models generated summaries with more \textit{redundancies} compared to golden reference summaries, as models may easily lost focus on massive information; (ii) once we introduced certain conversation structures such as topic-view and stage-view, models behaved better in terms of \textit{redundancy} and \textit{incorrect reasoning}, which indicated that the structured views could help models to better understand the conversations; (iii) our multi-view models which combined both stage-view and topic-view made the least number of errors compared to all single view models, suggesting the effectiveness of combining different views for conversation summarization.

\end{document}